\documentclass{article}
\usepackage{amsmath,amssymb,xspace,amsthm,Tabbing,bm}
\usepackage{color,mathtools}
\usepackage{amsthm}
\usepackage{enumitem}
\usepackage{listings}
\usepackage{multirow}
\usepackage{fullpage}
\usepackage{bbm}
\usepackage[export]{adjustbox}
\usepackage{wrapfig}
\usepackage{tikz}
\usepackage{wasysym}

\usepackage{calrsfs}

\DeclareMathAlphabet{\pazocal}{OMS}{zplm}{m}{n}
\usepackage{xstring}

\newcommand\transpose[1]{
  \begingroup
  \def\ADJa{\mkern-3mu}
  \noexpandarg
  #1^{
    \IfEndWith{#1}{A}{\ADJa}{}
    \IfEndWith{#1}{L}{\ADJa}{}
    \IfEndWith{#1}{u}{\ADJa}{}
    \IfEndWith{#1}{w}{\ADJa}{}
    \top
  }
  \endgroup
}

\usepackage{times}
\usepackage{graphicx} 

\usepackage{natbib}

\usepackage{graphicx}
\usepackage{caption}
\usepackage{subcaption}

\usepackage{algorithm}
\usepackage{algorithmic}

\providecommand{\nor}[1]{\left\lVert {#1} \right\rVert}

\providecommand{\scal}[2]{\left\langle{#1},{#2}\right\rangle}

\providecommand{\scalT}[2]{\left\langle{#1},{#2}\right\rangle}

\providecommand{\scal}[2]{\left\langle{#1},{#2}\right\rangle}

\usepackage{hyperref}

\usepackage[accepted]{icml2017}

\icmltitlerunning{McGan: Mean and Covariance Feature Matching GAN}

\begin{document} 

\twocolumn[
\icmltitle{McGan: Mean and Covariance Feature Matching GAN}



\icmlsetsymbol{equal}{*}

\begin{icmlauthorlist}
\icmlauthor{Youssef  Mroueh}{equal,aif,to}
\icmlauthor{Tom Sercu}{equal,aif,to}
\icmlauthor{Vaibhava Goel}{to}
\end{icmlauthorlist}

\icmlaffiliation{aif}{AI Foundations.
IBM T.J. Watson Research Center, NY, USA}
\icmlaffiliation{to}{Watson Multimodal Algorithms and Engines Group.
IBM T.J. Watson Research Center, NY, USA}

\icmlcorrespondingauthor{Youssef Mroueh}{mroueh@us.ibm.com}

\icmlkeywords{Generative Adversarial Networks, Integral probability metrics, McGan, Covariance Feature Matching}

\vskip 0.3in
]



\printAffiliationsAndNotice{\icmlEqualContribution} 

\begin{abstract} 
We introduce new families of  Integral Probability Metrics (IPM) for training Generative Adversarial Networks (GAN).
Our IPMs are based on matching  statistics of distributions  embedded in a finite dimensional feature space. Mean and covariance feature matching IPMs allow for stable training of GANs, which we will call McGan. McGan minimizes a meaningful loss between distributions. 

\end{abstract}

\section{Introduction}
Unsupervised learning of distributions is an important problem, in which we aim to learn underlying features that unveil the hidden the structure in the data.
The classic approach to learning distributions is by explicitly parametrizing the data likelihood and fitting this model by maximizing the likelihood of the real data.
An alternative recent approach is to learn a generative model of the data without explicit parametrization of the likelihood. 
Variational Auto-Encoders (VAE) \cite{VAE} and Generative Adversarial Networks (GAN) \cite{GANoriginal} fall under this category. 

We focus on the GAN approach.
In a nutshell GANs learn a generator of the data via a min-max game between the generator and a discriminator, which learns to distinguish between ``real'' and ``fake'' samples.
In this work we focus on the objective function that is being minimized between the learned generator distribution $\mathbb{P}_{\theta}$ and the real data distribution $\mathbb{P}_r$.

The original work of \cite{GANoriginal} showed that in GAN  this objective is the Jensen-Shannon divergence. \cite{FGAN} showed that other $\varphi$-divergences can be successfully used. The Maximum Mean Discrepancy objective (MMD) for GAN training was proposed in \cite{mmdGAN1,mmdGAN2}. As shown empirically in \cite{openAiGan}, one can train the GAN discriminator using the objective of \cite{GANoriginal} while training the generator using mean feature matching. An energy based objective for GANs was also developed recently \cite{EBGAN}. Finally, closely related  to our paper, the recent work \emph{Wasserstein GAN} (WGAN) of \cite{WGAN} proposed to use the Earth Moving distance (EM) as an objective for training GANs. Furthermore \cite{WGAN} show that the EM objective has many advantages as the loss function correlates with the quality of the generated samples and the \emph{mode dropping} problem is reduced in WGAN.    

In this paper, inspired by the MMD distance and the kernel mean embedding of distributions \cite{KernelMeanEmbedding} we propose to embed distributions in a finite dimensional feature space and to match them based on their \textbf{mean and covariance feature statistics}. Incorporating first and second order statistics has a better chance to  capture the various modes of the distribution. While mean feature matching was empirically used in \cite{openAiGan}, we show in this work that it is theoretically grounded: similarly to the EM distance in \cite{WGAN}, mean and covariance feature matching of two distributions can be written as a distance in the framework of Integral Probability Metrics (IPM) \cite{IPM}. 
To match the means, we can use any $\ell_q$ norm, hence we refer to mean matching IPM, as IPM$_{\mu,q}$. 
For matching covariances, in this paper we consider the Ky-Fan norm, which can be computed cheaply without explicitly constructing the full covariance matrices, and refer to the corresponding IPM as IPM$_{\Sigma}$.  

Our technical contributions can be summarized as follows:

\noindent a) We show in Section \ref{Sec:muIPM} that the $\ell_q$ mean feature matching  IPM$_{\mu,q}$ has two equivalent primal and dual formulations and can be used as an objective for GAN training in both formulations.

\noindent b) We show in Section \ref{Sec:equivalence} that the parametrization used in \emph{Wasserstein GAN} corresponds to $\ell_1$ mean feature matching GAN (IPM$_{\mu,1}$ GAN in our framework). 

\noindent c)~ We show in Section \ref{Sec:SigmaIPM} that the covariance feature matching IPM$_{\Sigma}$ admits also two dual formulations, and can be used as an objective for GAN training.

\noindent d)~ Similar to \emph{Wasserstein GAN}, we show that mean feature matching  and covariance matching GANs (McGan) are stable to train, have a reduced mode dropping and the IPM loss correlates with the quality of the generated samples.

\section{Integral Probability Metrics}\label{sec:IPM}
We define in this Section IPMs as a distance between distribution. Intuitively each IPM finds  a ``critic" $f$ \cite{WGAN} which maximally discriminates between the distributions. 
\subsection{IPM Definition}
Consider a compact space $\pazocal{X}$ in $\mathbb{R}^{d}$. Let $\mathcal{F}$ be a set of measurable and bounded real valued  functions on $\pazocal{X}$. Let $\mathcal{P}(\pazocal{X})$ be the set of measurable probability distributions on $\pazocal{X}$. Given two probability distributions $\mathbb{P},\mathbb{Q} \in \mathcal{P}({\pazocal{X}})$, the Integral probability metric (IPM) indexed by the function space $\mathcal{F}$ is defined as follows \cite{IPM}:
$$d_{\mathcal{F}}(\mathbb{P},\mathbb{Q})= \sup_{f \in \mathcal{F}} \Big | \underset{x\sim \mathbb{P}}{\mathbb{E}} f(x) -\underset{x\sim \mathbb{Q}}{\mathbb{E}}f(x)\Big|. $$
In this paper we are interested in symmetric function spaces $\mathcal{F}$, i.e $\forall f \in \mathcal{F}, -f \in \mathcal{F}$, hence we can write the IPM in that case without the absolute value:
\begin{equation}
d_{\mathcal{F}}(\mathbb{P},\mathbb{Q})= \sup_{f \in \mathcal{F}}\Big\{  \underset{x\sim \mathbb{P}}{\mathbb{E}} f(x) -\underset{x\sim \mathbb{Q}}{\mathbb{E}}f(x)\Big\}.
\label{eq:IPM}
\end{equation} 
It is easy to see that $d_{\mathcal{F}}$ defines a pseudo-metric over $\mathcal{P}(X)$. ($d_{\mathcal{F}}$ non-negative, symmetric and satisfies the triangle inequality. A pseudo metric means that $d_{\mathcal{F}}(\mathbb{P},\mathbb{P})=0$ but $d_{\mathcal{F}}(\mathbb{P},\mathbb{Q})=0$ does not necessarily imply $\mathbb{P}=\mathbb{Q}$).

By choosing $\mathcal{F}$ appropriately \cite{IPMemp,IPMDivergence}, various distances between probability measures can be defined. In the next subsection following \cite{WGAN,mmdGAN1,mmdGAN2} we show how to use IPM to learn generative models of distributions, we then specify a special set of functions $\mathcal{F}$ that makes the learning tractable.
\subsection{Learning Generative Models with IPM }
In order to learn a generative model of  a distribution $\mathbb{P}_r\in \mathcal{P}(\pazocal{X})$, we learn a function $$g_{\theta}: \pazocal{Z}\subset \mathbb{R}^{n_z}\to \pazocal{X},$$
such that for $z\sim p_{z}$, the distribution of $g_{\theta}(z)$ is close to the real data distribution $\mathbb{P}_{r}$, where $p_{z}$ is a fixed distribution on $\pazocal{Z}$ (for instance $z\sim \mathcal{N}(0,I_{n_z})$). 
Let $\mathbb{P}_{\theta}$ be the distribution of $g_{\theta}(z),z\sim p_{z}$. Using an IPM indexed by a function class $\mathcal{F}$ we shall solve therefore the following problem:
\begin{equation}
\min_{g_{\theta}} d_{\mathcal{F}}(\mathbb{P}_r,\mathbb{P}_{\theta}) 
\label{eq:IPMGAN}
\end{equation}
Hence this amounts to solving the following min-max problem:
$$\min_{g_{\theta}} \sup_{f\in \mathcal{F}}\underset{x\sim \mathbb{P}_r}{\mathbb{E}}f(x) -\underset{z \sim p_{z}}{\mathbb{E}}f(g_{\theta}(z))  $$
Given samples $\{x_i ,1\dots N\}$ from $\mathbb{P}_r$ and samples $\{z_i ,1\dots M\}$ from $p_{z}$ we shall solve the following empirical problem:
$$\min_{g_{\theta}} \sup_{f\in \mathcal{F}} \frac{1}{N} \sum_{i=1}^N f(x_i) - \frac{1}{M}\sum_{j=1}^M f(g_{\theta}(z_j)), $$
in the following we consider for simplicity $M=N$.
\begin{figure*}[ht!]
\centering
 \includegraphics[height=3in,width=5in]{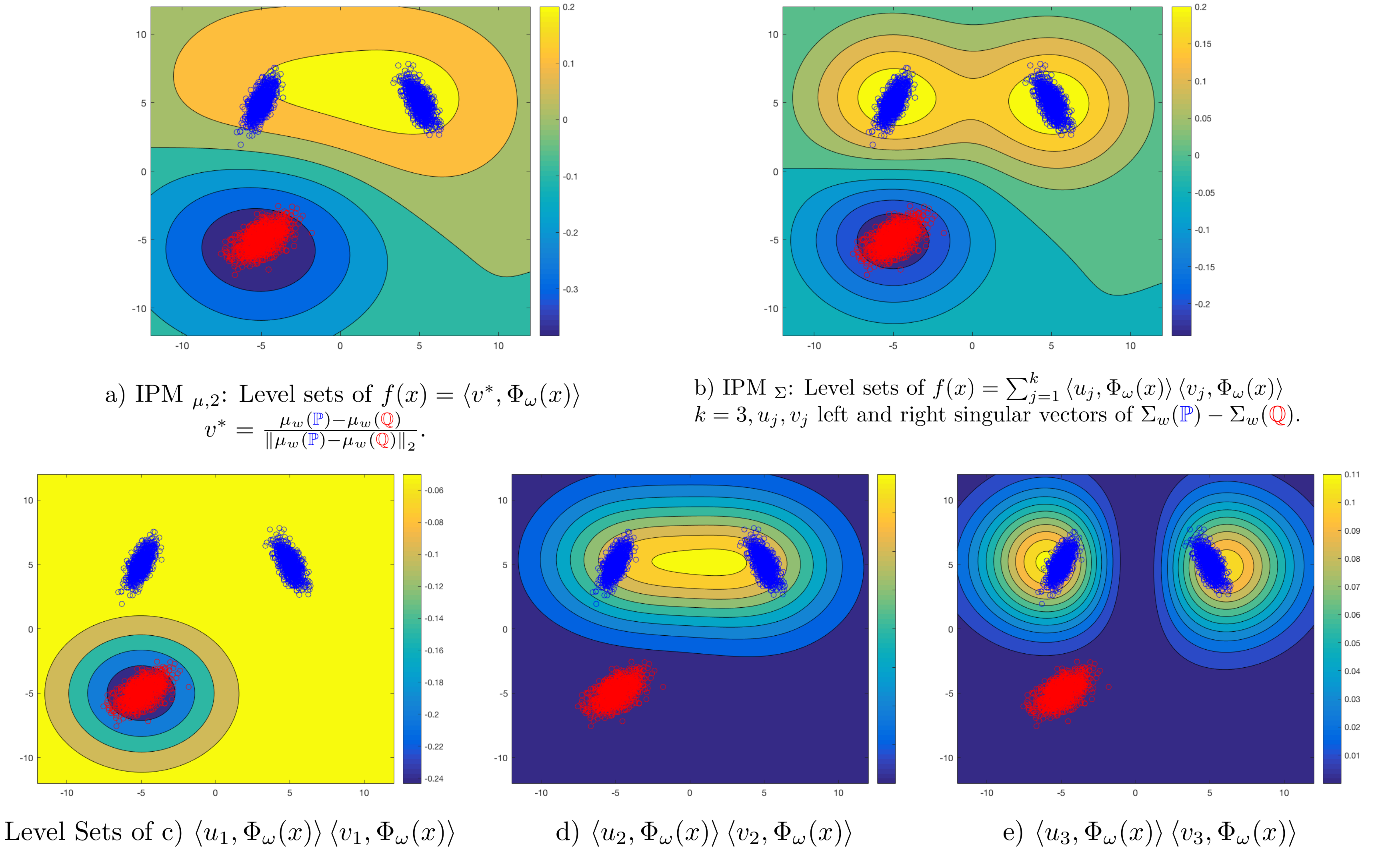}
 \caption{
   Motivating example on synthetic data in 2D, showing how different components in covariance matching can target different regions of the input space.
   Mean matching (a) is not able to capture the two modes of the bimodal ``real'' distribution $\textcolor{blue}{\mathbb{P}}$ and assigns higher values to one of the modes.
   Covariance matching (b) is composed of the sum of three components (c)+(d)+(e), corresponding to the top three ``critic directions''.
   Interestingly, the first direction (c) focuses on the ``fake'' data $\textcolor{red}{\mathbb{Q}}$, 
   the second direction (d) focuses on the ``real'' data,
   while the third direction (e) is mode selective.
   This suggests that using covariance matching would help reduce mode dropping in GAN. 
   In this toy example $\Phi_{\omega}$ is a fixed random Fourier feature map \cite{rf} of a Gaussian kernel (i.e. a finite dimensional approximation).
 }

\label{fig:levelset}
\end{figure*}
\section{Mean Feature Matching GAN}\label{Sec:muIPM}
In this Section we introduce a class of functions $\mathcal{F}$ having the form $\scalT{v}{\Phi_{\omega}(x)}$, where  vector $v \in \mathbb{R}^m$ and  $\Phi_{\omega}: \pazocal{X}\to \mathbb{R}^m$ a non linear feature map (typically parametrized by a neural network). We show in this Section that the IPM defined by this function class corresponds to the distance between the mean of the distribution in  the $\Phi_{\omega}$ space.  
\subsection{IPM$_{\mu,q}$: Mean Matching IPM}

More formally consider the following function space: 
\vskip -2em
\begin{align*}
&\mathcal{F}_{v,\omega,p}=\{ f(x)= \langle v,\Phi_{\omega}(x)\rangle \Big | v \in \mathbb{R}^{m}, \nor{v}_p\leq 1,\\
& \Phi_{\omega}: \pazocal{X} \to \mathbb{R}^{m}, \omega \in \Omega \},
 \end{align*}
where $\nor{.}_{p}$ is the $\ell_{p}$ norm.
$\mathcal{F}_{v,\omega,p}$ is the space of bounded linear functions defined in the non linear feature space induced by the parametric feature map  $\Phi_{\omega}$. $\Phi_{\omega}$ is typically a multi-layer neural network. The parameter space $\Omega$ is chosen so that  the function space $ \mathcal{F}$ is bounded. Note that for a given $\omega$, $\mathcal{F}_{v,\omega,p}$  is a finite dimensional Hilbert space.

We recall here  simple definitions on dual norms that will be necessary for the analysis in this Section.
Let $p,q \in [1,\infty]$, such that $\frac{1}{p}+\frac{1}{q}=1$. 
By duality of norms we have:
$\nor{x}_{q}=\max_{v, \nor{v}_{p}\leq 1} \scalT{v}{x}$ and the  Holder inequality:
$\Big|\scalT{x}{y}\Big| \leq \nor{x}_{p}\nor{y}_{q}$.

From Holder inequality  we obtain the following bound:
$$\Big|f(x)\Big|=\Big|\scalT{v}{\Phi_{\omega}x}\Big|\leq \nor{v}_p\nor{\Phi_{\omega}(x)}_q\leq  \nor{\Phi_{\omega}(x)}_q.$$ 
To ensure that $f$ is bounded, it is enough to consider $\Omega$ such that $\nor{\Phi_{\omega}(x)}_{q}\leq B, \forall~ x \in \pazocal{X}$. Given that the space $\pazocal{X}$ is bounded it is sufficient to control the norm of the weights and biases of the neural network $\Phi_{\omega}$ by regularizing the $\ell_{\infty}$ (clamping) or $\ell_2$ norms (weight decay) to ensure the boundedness of  $\mathcal{F}_{v,\omega,p}$.

Now that we ensured the boundedness of $\mathcal{F}_{v,\omega,p}$ , we  look at its corresponding IPM:
 \begin{eqnarray*}
& &d_{\mathcal{F}_{v,\omega,p}}(\mathbb{P},\mathbb{Q})=\sup_{f \in \mathcal{F}_{v,\omega,p}}  \underset{x\sim \mathbb{P}}{\mathbb{E}} f(x) -\underset{x\sim \mathbb{Q}}{\mathbb{E}}f(x) \\
&=&\max_{\omega \in \Omega, v,||v||_{p}\leq1} \scalT{ v}{ \underset{x\sim \mathbb{P}}{\mathbb{E}}\Phi_{\omega}(x)- \underset{x\sim \mathbb{Q}}{\mathbb{E}}  \Phi_{\omega}(x) }\\
&=& \max_{\omega \in \Omega}\Big[ \max_{ v,||v||_{p}\leq1} \scalT{ v}{ \underset{x\sim \mathbb{P}}{\mathbb{E}}\Phi_{\omega}(x)- \underset{x\sim \mathbb{Q}}{\mathbb{E}}  \Phi_{\omega}(x) }\Big]\\
&=& \max_{\omega \in \Omega} \nor{\mu_{\omega}(\mathbb{P})-\mu_{\omega}(\mathbb{Q})}_q,
\end{eqnarray*}
where we used the linearity of the function class and expectation in the first equality and the definition of the dual norm $\nor{.}_{q}$ in the last equality and our
definition of  the mean feature embedding of  a distribution $\mathbb{P}\in \mathcal{P}(\pazocal{X})$:
$$\mu_{\omega}(\mathbb{P})= \underset{x\sim \mathbb{P}}{\mathbb{E}}\Big[ \Phi_{\omega}(x)\Big] \in \mathbb{R}^{m}.$$

We see that the IPM indexed by $\mathcal{F}_{v,\omega,p}$, corresponds to the Maximum mean feature Discrepancy between the two distributions. Where the maximum is taken over the parameter set $\Omega$, and the discrepancy is measured in the $\ell_{q}$ sense between the mean feature embedding of $\mathbb{P}$ and $\mathbb{Q}$. In other words this IPM is equal to the worst case $\ell_{q}$ distance between mean feature embeddings of distributions. We refer in what follows to $d_{\mathcal{F}_{v,\omega,p}}$ as IPM$_{\mu,q}$.

\subsection{Mean Feature Matching GAN}
We turn now to the problem of learning generative models with IPM$_{\mu,q}$.
Setting $\mathcal{F}$ to $\mathcal{F}_{v,\omega,p}$ in Equation \eqref{eq:IPMGAN} yields to the following min-max problem for learning generative models:
\begin{equation}
\min_{g_\theta} \max_{\omega \in \Omega} \max_{v,||v||_{p}\leq 1} \mathcal{L}_{\mu}(v,\omega,\theta), 
\label{eq:GAN-v-mean}
\end{equation}
where
 \vskip -0.3in
$$ \mathcal{L}_{\mu}(v,\omega,\theta)=\scalT{ v}{\underset{x\sim \mathbb{P}_r}{\mathbb{E}}\Phi_{\omega}(x)- \underset{z \sim p_{z}}{\mathbb{E}}  \Phi_{\omega}(g_{\theta}(z)) },$$
or equivalently using the dual norm:
\begin{equation}
\min_{g_{\theta}}\max_{\omega \in \Omega} \nor{\mu_{\omega}(\mathbb{P}_r)-\mu_{\omega}(\mathbb{P}_{\theta})}_q ,
\label{eq:GANmean}
\end{equation}
where $\mu_{\omega}(\mathbb{P}_{\theta})=\underset{z \sim p_{z}}{\mathbb{E}} \Phi_{\omega}(g_{\theta}(z))$.

We refer to formulations \eqref{eq:GAN-v-mean} and \eqref{eq:GANmean} as primal and dual formulation respectively.

The dual formulation in Equation \eqref{eq:GANmean} has a simple interpretation as an adversarial learning game: while the feature space $\Phi_{\omega}$ tries to  map the mean feature embeddings of the \emph{real} distribution $\mathbb{P}_r$ and the \emph{fake}  distribution $\mathbb{P}_{\theta}$ to be far apart (maximize the $\ell_q$ distance between the mean embeddings), the generator $g_{\theta}$ tries to put them close one to another. Hence we refer to this IPM as \emph{mean matching} IPM.

We devise empirical estimates of both formulations in Equations \eqref{eq:GAN-v-mean} and \eqref{eq:GANmean}, given samples $\{x_i,i=1\dots N\}$ from $\mathbb{P}_r$, and $\{z_i,i=1\dots N\}$ from $p_z$. The primal formulation \eqref{eq:GAN-v-mean} is more amenable to stochastic gradient descent since the expectation operation appears in a linear way in the cost function of Equation \eqref{eq:GAN-v-mean}, while it is non linear in the cost function of the dual formulation \eqref{eq:GANmean} (inside the norm). We give here the empirical estimate of the primal  formulation by giving empirical estimates  $\hat{\mathcal{L}}_{\mu}(v,\omega,\theta)$ of the primal cost function:
\begin{equation*}
(P_{\mu}):\min_{g_\theta} ~~~\underset{\mathclap{\substack{\omega \in \Omega\\v,||v||_{p}\leq1}}}{\max}~~~\scalT{ v}{ \frac{1}{N}\sum_{i=1}^N\Phi_{\omega}(x_i)- \frac{1}{N} \sum_{i=1}^N \Phi_{\omega}(g_{\theta}(z_i)) } 
\label{eq:PmeanMatchingIMPGAN}
\end{equation*}
An empirical estimate of the dual formulation can be also given as follows:
\begin{equation*}
(D_{\mu}):\min_{g_\theta} ~~~\underset{\mathclap{\substack{\omega \in \Omega}}}{\max}~~~ \nor{\frac{1}{N}\sum_{i=1}^N\Phi_{\omega}(x_i)-\frac{1}{N} \sum_{i=1}^N \Phi_{\omega}(g_{\theta}(z_i)) }_{q}
\label{eq:DmeanMatchingIMPGAN}
\end{equation*}
In what follows we refer to the problem given in (P$_{\mu}$) and (D$_{\mu}$) as \emph{$\ell_{q}$ Mean Feature Matching} GAN.
Note that while (P$_{\mu}$) does not need real samples for optimizing the generator, (D$_{\mu}$) does need samples from real and fake. 
Furthermore we will need a large minibatch of real data in order to get a good estimate of the expectation. This makes the primal formulation more appealing computationally.


\subsection{Related Work} \label{Sec:equivalence}
We show in this Section that several previous works on GAN, can be written within the $\ell_q$ mean feature matching IPM (IPM$_{\mu,q}$) minimization framework:

a) \emph{Wasserstein GAN} (WGAN): \cite{WGAN} recently introduced Wasserstein GAN. While the main motivation of this paper is to consider the IPM indexed by Lipchitz functions on $\pazocal{X}$, we show that the particular parametrization considered in \cite{WGAN}  corresponds to a  mean feature matching IPM.\\
Indeed \cite{WGAN} consider the function set parametrized by a convolutional neural network with a linear output layer and weight clipping.
Written in our notation, the last linear layer corresponds to $v$, and the convolutional neural network below corresponds to $\Phi_{\omega}$.
Since $v$ and $\omega$ are simultaneously clamped, this corresponds to restricting $v$ to be in the $\ell_\infty$ unit ball, and to define in $\Omega$ constraints on the $\ell_{\infty}$ norms of $\omega$. In other words \cite{WGAN} consider functions in $\mathcal{F}_{v,\omega,p}$, where $p=\infty$ . 
Setting $p=\infty$ in Equation \eqref{eq:GAN-v-mean}, and $q=1$ in Equation \eqref{eq:GANmean}, we see that in WGAN we are minimizing $d_{\mathcal{F}_{v,\omega,\infty}}$, that corresponds to $\ell_1$ mean feature matching GAN.
 
b) \emph{MMD GAN}: Let $\mathcal{H}$ be a Reproducing Kernel Hilbert Space (RKHS) with $k$ its reproducing kernel. For any valid PSD kernel $k$ there exists an \emph{infinite dimensional feature} map $\Phi: \pazocal{X}\to \mathcal{H}$ such that: $k(x,y)=\langle \Phi(x),\Phi(y)\rangle_{\mathcal{H}}$. For an RKHS $\Phi$ is noted usually $k(x,.)$ and satisfies the reproducing proprety:
$$ f(x)=\langle f,\Phi(x)\rangle_{\mathcal{H}}, \text{ for all  } f \in \mathcal{H}. $$
 Setting $\mathcal{F}=\big\{ f \big| \left|\left| f \right|\right|_{\mathcal{H}} \leq 1\big\}$ in Equation \eqref{eq:IPM} the IPM  $d_{\mathcal{F}}$ has a simple expression:
\begin{eqnarray}
d_{\mathcal{F}}(\mathbb{P},\mathbb{Q})&=& \sup_{f, ||f||_{\mathcal{H} }\leq 1}\Big\{\Big \langle f,  \underset{x\sim \mathbb{P}}{\mathbb{E}} \Phi (x) -\underset{x\sim \mathbb{Q}}{\mathbb{E}}\Phi(x)\Big \rangle\Big\} \nonumber\\
&=& \Big | \Big |\mu(\mathbb{P}) -\mu(\mathbb{Q})\Big|\Big |_{\mathcal{H}},
\label{eq:mmd}
\end{eqnarray}
where $\mu(\mathbb{P})=\underset{x\sim \mathbb{P}}{\mathbb{E}} \Phi (x) \in \mathcal{H}$ is the so called kernel mean embedding \cite{KernelMeanEmbedding}. $d_{\mathcal{F}}$ in this case is the so called Maximum kernel Mean Discrepancy (MMD) \cite{MMD} . Using the reproducing property MMD has a closed form in term of the kernel $k$. Note that IPM$_{\mu,2}$ is a special case of MMD when the feature map is finite dimensional, with the main difference that the feature map is fixed in case of MMD and learned in the case of IPM$_{\mu,2}$.  
\cite{mmdGAN1,mmdGAN2} showed that GANs can be learned using MMD with a fixed gaussian kernel.

c) \emph{Improved GAN}: Building on the pioneering work of \cite{GANoriginal}, \cite{openAiGan} suggested to learn the discriminator with the binary cross entropy criterium of GAN while  learning the generator with $\ell_2$ mean feature matching. The main difference of our IPM$_{\mu,2}$ GAN is that both ``discriminator" and ``generator" are learned using the mean feature matching criterium, with additional constraints on $\Phi_{\omega}$.

 \section{Covariance Feature Matching GAN}
 \subsection{IPM$_{\Sigma}$: Covariance Matching IPM}
 As follows from our discussion of mean matching IPM comparing two distributions amounts to comparing  a first order statistics, the mean of their feature embeddings.
Here we ask the question how to incorporate second order statistics, i.e covariance information of feature embeddings.
  
In this Section we will provide a function space $\mathcal{F}$  such that the IPM in Equation \eqref{eq:IPM} captures second order information. 
Intuitively a distribution of points represented in a feature space can be approximately captured by its mean and its covariance.
Commonly in unsupervised learning, this covariance is approximated by its first $k$ principal components (PCA directions), which capture the directions of maximal variance in the data.
Similarly, the metric we define in this Section will find $k$ directions that maximize the discrimination between the two covariances.
Adding second order information would enrich the discrimination power of the feature space (See Figure \ref{fig:levelset}).

 
This intuition motivates the following function space of bilinear functions  in $\Phi_{\omega}$ :
\vspace{-0.1 in}
\begin{align*}
&\mathcal{F}_{U,V,\omega}=\{f(x)=\sum_{j=1}^k \scalT{u_j}{\Phi_{\omega}(x)} \scalT{v_j}{\Phi_{\omega}(x)} \\
& \{u_j\},\{v_j\} \in \mathbb{R}^{m} \text{ orthonormal } j=1\dots k,  \omega \in \Omega  \}.
\end{align*}
Note that the set $\mathcal{F}_{U,V,\omega}$ is symmetric and hence the IPM indexed by this set  (Equation \eqref{eq:IPM}) is well defined. It is easy to see that $\mathcal{F}_{U,V,\omega}$ can be written as:
\begin{align*}
&\mathcal{F}_{U,V,\omega}=\{ f(x)= \scalT{U^{\top}\Phi_{\omega}(x)}{V^{\top}\Phi_{\omega}(x)} ) \Big|\\
& U,V \in \mathbb{R}^{m\times k}, U^{\top}U=I_{k},V^{\top}V=I_{k},  \omega \in \Omega \}
\end{align*}
the parameter set $\Omega$ is such that the function space remains bounded. 
Let $$\Sigma_{\omega}(\mathbb{P})= \underset{x\sim \mathbb{P}}{\mathbb{E}}\Phi_{\omega}(x)\Phi_{\omega}(x)^{\top},$$
be the uncentered feature covariance embedding of $\mathbb{P}$. It is easy to see that $\underset{x\sim \mathbb{P}}{\mathbb{E}}f(x)$ can be written in terms of $U,V,$ and $\Sigma_{\omega}(\mathbb{P})$:
$$\underset{x\sim \mathbb{P}}{\mathbb{E}}f(x)=\underset{x\sim \mathbb{P}}{\mathbb{E}}\scal{U^{\top}\Phi(x)}{V^{\top}\Phi(x)}=Trace(U^{\top}\Sigma_{\omega}(\mathbb{P})V).$$
\vskip -0.14 in
 For a matrix $A \in \mathbb{R}^{m\times m}$, we note by $\sigma_j(A)$ the singular value of A, $j=1\dots m$ in descending order.
The 1-schatten norm or the nuclear norm is defined as the sum of singular values, $\nor{A}_{*}=\sum_{j=1}^m \sigma_j$. We note by $[A]_{k}$ the k-th rank approximation of $A$.
We note $\pazocal{O}_{m,k}=\{M\in \mathbb{R}^{m\times k} | M^{\top}M=I_{k} \}$.
\noindent Consider the IPM induced by this function set. Let $\mathbb{P},\mathbb{Q}\in \mathcal{P}(\pazocal{X}) $ we have:
\begin{align*}
& d_{\mathcal{F}_{U,V,\omega}}(\mathbb{P},\mathbb{Q})=\sup_{f \in\mathcal{F}_{U,V,\omega}} \underset{x\sim \mathbb{P}}{\mathbb{E}}f(x) - \underset{x\sim \mathbb{Q}}{\mathbb{E}}f(x)  \\
&=\underset{\mathclap{\substack{\omega\in \Omega\\ ~U,V \in \pazocal{O}_{m,k}}}}{\max }~ \underset{x\sim \mathbb{P}}{\mathbb{E}}f(x)-\underset{x\sim \mathbb{Q}}{\mathbb{E}}f(x) \\
&= \underset{\omega \in \Omega}{\max}~~~~ \underset{\mathclap{\substack{U,V \in \pazocal{O}_{m,k} }}}{\max }~~Trace\left[U^{\top}( \Sigma_{\omega}(\mathbb{P})-\Sigma_{\omega}(\mathbb{Q}) )V\right]\\
&= \max_{\omega \in \Omega}  \sum_{j=1}^k \sigma_{j}\left(\Sigma_{\omega}(\mathbb{P})-\Sigma_{\omega}(\mathbb{Q})\right)\\
&= \max_{\omega \in \Omega} \nor{\left[\Sigma_{\omega}(\mathbb{P})- \Sigma_{\omega}(\mathbb{Q})\right]_{k}}_{*},
\end{align*}
where we used the variational definition of singular values and the definition of the nuclear norm.
Note that $U,V$ are the left and right singular vectors of $\Sigma_{\omega}(\mathbb{P})- \Sigma_{\omega}(\mathbb{Q})$.
Hence $d_{\mathcal{F}_{U,V,\omega}}$ measures the worst case distance between the covariance feature embeddings of the two distributions, this distance is measured with the  \emph{Ky Fan $k$-norm} (nuclear norm of truncated covariance difference). Hence we call this IPM covariance matching IPM, IPM$_{\Sigma}$.

\subsection{Covariance Matching GAN}\label{Sec:SigmaIPM}
Turning now to the problem of learning a generative model $g_{\theta}$ of $\mathbb{P}_r\in \mathcal{P}(\pazocal{X})$ using  IPM$_{\Sigma}$ we shall solve:
$$\min_{g_{\theta}}d_{\mathcal{F}_{U,V,\omega}}(\mathbb{P}_r,\mathbb{P}_{\theta}),$$
this has the following primal formulation:
\begin{equation}
\min_{g_{\theta}}~~~~~~~~~\underset{\mathclap{\substack{ \omega \in \Omega, U,V \in \pazocal{O}_{m,k}}}}{\max }~~~~~~\mathcal{L}_{\sigma}(U,V,\omega,\theta)  ,
\label{eq:GANsubMatch}
\end{equation} 
\vskip -0.2in 
\begin{align*}
&\text{where } \mathcal{L}_{\sigma}(U,V,\omega,\theta)=\underset{x\sim \mathbb{P}_r}{\mathbb{E}}\scalT{\transpose{U}\Phi_{\omega}(x)}{\transpose{V}\Phi_{\omega}(x))}\\
&-\underset{z\sim p_{z}}{\mathbb{E}}\scalT{\transpose{U}\Phi_{\omega}(g_{\theta}(z))}{\transpose{V}\Phi_{\omega}(g_{\theta}(z))},
 \end{align*}
or equivalently the following dual formulation: 
 \begin{equation}
\min_{g_{\theta}} \max_{\omega \in \Omega} \nor{[\Sigma_{\omega}(\mathbb{P}_r)- \Sigma_{\omega}(\mathbb{P_{\theta}})]_{k}}_*, 
\label{eq:GANcovMatch}
 \end{equation}
 where $ \Sigma_{\omega}(\mathbb{P_{\theta}})=\mathbb{E}_{z\sim p_{z}}\Phi_{\omega}(g_{\theta}(z))\Phi_{\omega}(g_{\theta}(z))^{\top}$.
 
The dual formulation in Equation \eqref{eq:GANcovMatch} shows that learning generative models with IPM$_{\Sigma}$, consists in an adversarial game between the feature map and the generator, when the feature maps tries to maximize the distance between the  feature covariance  embeddings of the distributions, the generator tries to minimize this distance. Hence we call learning with IPM$_{\Sigma}$, covariance matching GAN.
 
\noindent We give here an empirical estimate of the primal formulation in  Equation \eqref{eq:GANsubMatch} which is amenable to stochastic gradient. The dual requires nuclear norm minimization and is more involved.
 Given $\{x_i, x_i\sim \mathbb{P}_r\}$, and $\{z_j, z_j \sim p_{z}\}$, the \emph{covariance matching GAN} can be written as follows:
\begin{equation}
\min_{g_{\theta}}~~~~~~~~~\underset{\mathclap{\substack{\omega \in \Omega,U,V \in \pazocal{O}_{m,k}}}}{\max }~~~~~~\hat{\mathcal{L}}_{\sigma}(U,V,\omega,\theta)  ,
\label{eq:GANsubMatchemp}
\end{equation} 
\begin{alignat*}{3}
&\text{where }\hat{\mathcal{L}}_{\sigma}(U,V,\omega,\theta) =\frac{1}{N} \sum_{i=1}^N \scalT{\transpose{U}\Phi_{\omega}(x_i)}{\transpose{V}\Phi_{\omega}(x_i)}\\ 
 &-\frac{1}{N} \sum_{j=1}^N \scalT{\transpose{U} \Phi_{\omega}(g_{\theta}(z_j))}{\transpose{V}\Phi_{\omega}(g_{\theta}(z_j))}.
\end{alignat*}

\subsection{Mean and Covariance Matching GAN}
In order to match first and second order statistics we propose the following simple extension:
\begin{equation*}
\min_{g_{\theta}}\underset{\substack{\omega \in \Omega,v,||v||_{p}\leq 1\\U,V \in \pazocal{O}_{m,k}}}{\max} \mathcal{L}_{\mu}(v,\omega,\theta)+\mathcal{L}_{\sigma}(U,V,\omega,\theta),
\label{eq:meancovP}
\end{equation*}
that has a simple dual adversarial game interpretation
\begin{equation*}
\min_{g_{\theta}}\underset{\omega \in \Omega}{\max} \nor{\mu_{\omega}(\mathbb{P})-\mu_{\omega}(\mathbb{P}_{\theta})}_{q}+ \nor{[\Sigma_{\omega}(\mathbb{P}_r)- \Sigma_{\omega}(\mathbb{P_{\theta}})]_{k}}_*,
\end{equation*}
where the discriminator finds a feature space that discriminates between means and variances of real and fake, and the generator tries to match the real statistics.
We can also give empirical estimates of the primal formulation similar to expressions given in the paper.

\section{Algorithms}\label{Sec:algo}
We present in this Section our algorithms for mean and covariance feature matching GAN (McGan) with IPM$_{\mu,q}$ and IPM$_{\Sigma}$.

\noindent \textbf{Mean Matching GAN.} \emph{Primal P$_{\mu}$:} We give in Algorithm \ref{alg:P_mu} an algorithm for solving the primal IPM$_{\mu,q}$ GAN (P$_{\mu}$). Algorithm \ref{alg:P_mu} is adapted from \cite{WGAN} and corresponds to their algorithm for $p=\infty$. The main difference is that we allow projection of $v$ on different $\ell_{p}$ balls, and we maintain the clipping of $\omega$ to ensure boundedness of $\Phi_{\omega}$. For example for  $p=2$, $\text{proj}_{B_{\ell_2}}(v)=\min(1,\frac{1}{\nor{v}_2})v$. For $p=\infty$ we obtain the same clipping in \cite{WGAN} $\text{proj}_{B_{\ell_\infty}}(v)=\text{clip}(v,-c,c)$ for $c=1$.

\emph{Dual D$_{\mu}$:} We give in Algorithm \ref{alg:D_mu} an algorithm for solving the dual formulation IPM$_{\mu,q}$ GAN (D$_{\mu}$). As mentioned earlier we need samples from ``real" and ``fake" for training both generator and  the ``critic" feature space.

\noindent \textbf{Covariance Matching GAN.} \emph{Primal P$_{\Sigma}$:} We give in Algorithm \ref{alg:P_Sigma} an algorithm for solving the primal of IPM$_{\Sigma}$ GAN (Equation \eqref{eq:GANsubMatchemp}). The algorithm performs a stochastic gradient  ascent on $(\omega,U,V)$ and a descent on $\theta$.  We maintain clipping on $\omega$ to ensure boundedness of $\Phi_{\omega}$, and perform a QR retraction on the Stiefel manifold $\pazocal{O}_{m,k}$ \cite{retraction}, maintaining orthonormality of $U$ and $V$. 
\begin{algorithm}[ht!]
   \caption{Mean Matching GAN - Primal (P$_{\mu}$)}
   \label{alg:P_mu}
\begin{algorithmic}
   \STATE {\bfseries Input:} $p$ to define the ball of $v$ ,$\eta$ Learning rate, $n_c$ number of iterations for training the critic, $c$ clipping or weight decay parameter, N batch size
   \STATE {\bfseries Initialize} $v,\omega,\theta$
   \REPEAT
   \FOR{$j=1$ {\bfseries to} $n_c$}
   \STATE Sample  a minibatch $x_i,i=1\dots N, x_i \sim \mathbb{P}_r$ 
   \STATE Sample  a minibatch $z_i,i=1\dots N, z_i \sim p_z$ 
   \STATE $(g_{v},g_{\omega})\gets ( \nabla_{v} \hat{\mathcal{L}}_{\mu}(v,\omega,\theta),\nabla_{\omega} \hat{\mathcal{L}}_{\mu}(v,\omega,\theta)) $
   \STATE $(v,\omega )\gets (v,\omega) +\eta \text{ RMSProp }((v,\omega),(g_{v},g_{\omega}))$\\
   \COMMENT{Project $v$ on $\ell_p$ ball, $B_{\ell_p}=\{x,\nor{x}_{p}\leq 1\}$}
   \STATE $v \gets \text{proj}_{B_{\ell_p}}(v) $ 
   \STATE $\omega \gets \text{clip}(\omega,-c,c)$ \COMMENT{Ensure $\Phi_{\omega}$ is bounded}
   \ENDFOR
   \STATE Sample ${z_i,i=1\dots N , z_i \sim p_z}$
   \STATE $d_{\theta}\gets -\nabla_{\theta}\scalT{v}{\frac{1}{N}\sum_{i=1}^N \Phi_{\omega}(g_{\theta}(z_i))}$ 
   \STATE $\theta \gets \theta -\eta \text{ RMSProp }(\theta,d_{\theta})$
   \UNTIL{$\theta$  converges}
\end{algorithmic}
\end{algorithm}

\begin{algorithm}[ht!]
   \caption{Mean Matching GAN - Dual (D$_{\mu}$)}
   \label{alg:D_mu}
\begin{algorithmic}
   \STATE {\bfseries Input:} $q$ the matching $\ell_{q}$ norm ,$\eta$ Learning rate, $n_c$ number of iterations for training the critic, $c$ clipping or weight decay parameter, N batch size
   \STATE {\bfseries Initialize} $v,\omega,\theta$
   \REPEAT
   \FOR{$j=1$ {\bfseries to} $n_c$}
   \STATE Sample  a minibatch $x_i,i=1\dots N, x_i \sim \mathbb{P}_r$ 
   \STATE Sample  a minibatch $z_i,i=1\dots N, z_i \sim p_z$ 
   \STATE $\Delta_{\omega,\theta} \gets \frac{1}{N}\sum_{i=1}^N\Phi_{\omega}(x_i) -\frac{1}{N}\sum_{i=1}^N \Phi_{\omega}(g_{\theta}(z_i))$
   \STATE $g_{\omega}\gets \nabla_{\omega} \nor{\Delta_{\omega,\theta}}_{q} $
   \STATE $ \omega \gets \omega +\eta \text{ RMSProp }(\omega,g_{\omega})$
   \STATE $\omega \gets \text{clip}(\omega,-c,c)$ \COMMENT{Ensure $\Phi_{\omega}$ is bounded}
   \ENDFOR
   \STATE Sample ${z_i,i=1\dots N , z_i \sim p_z}$
   \STATE    Sample  $x_i,i=1\dots M, x_i \sim \mathbb{P}_r$ $(M>N)$
   \STATE  $\Delta_{\omega,\theta} \gets \frac{1}{M}\sum_{i=1}^M\Phi_{\omega}(x_i) -\frac{1}{N}\sum_{i=1}^N \Phi_{\omega}(g_{\theta}(z_i))$
   \STATE $d_{\theta}\gets \nabla_{\theta}\nor{\Delta_{\omega,\theta}}_{q}$ 
   \STATE $\theta \gets \theta -\eta \text{ RMSProp }(\theta,d_{\theta})$
   \UNTIL{$\theta$  converges}
\end{algorithmic}
\end{algorithm}

\begin{algorithm}[ht!]
   \caption{Covariance Matching GAN - Primal (P$_{\Sigma}$)}
   \label{alg:P_Sigma}
\begin{algorithmic}
   \STATE {\bfseries Input:} $k$ the number of components ,$\eta$ Learning rate, $n_c$ number of iterations for training the critic, $c$ clipping or weight decay parameter, N batch size
   \STATE {\bfseries Initialize} $U,V,\omega,\theta$
   \REPEAT
   \FOR{$j=1$ {\bfseries to} $n_c$}
   \STATE Sample  a minibatch $x_i,i=1\dots N, x_i \sim \mathbb{P}_r$ 
   \STATE Sample  a minibatch $z_i,i=1\dots N, z_i \sim p_z$ 
   \STATE $G\gets  (\nabla_{U},\nabla_{V},\nabla_{\omega}) \hat{\mathcal{L}}_{\sigma}(U,V,\omega,\theta)$
   \STATE $(U,V,\omega) \gets (U,V,\omega) +\eta \text{ RMSProp }((U,V,\omega),G)$
   \COMMENT{ Project U and V on the Stiefel manifold $\pazocal{O}_{m,k}$}
   \STATE $Q_u,R_u\gets QR(U) ~~ s_u \gets \text{sign}(\text{diag}(R_u))$
   \STATE $Q_v,R_v\gets QR(V) ~~ s_v \gets \text{sign}(\text{diag}(R_v))$
   \STATE $U \gets Q_u \text{Diag}(s_u)$
   \STATE $V \gets Q_v \text{Diag}(s_v)$\\
   \STATE $\omega \gets \text{clip}(\omega,-c,c)$ \COMMENT{Ensure $\Phi_{\omega}$ is bounded}
   \ENDFOR
   \STATE Sample ${z_i,i=1\dots N , z_i \sim p_z}$
   \STATE $d_{\theta}\gets -\nabla_{\theta} \frac{1}{N} \sum_{j=1}^N \scalT{U \Phi_{\omega}(g_{\theta}(z_j))}{V\Phi_{\omega}(g_{\theta}(z_j))}$ 
   \STATE $\theta \gets \theta -\eta \text{ RMSProp }(\theta,d_{\theta})$
   \UNTIL{$\theta$  converges}
\end{algorithmic}
\end{algorithm}

\section{Experiments}

We train McGan for image generation with both Mean Matching and Covariance Matching objectives. 
We show generated images on the labeled faces in the wild (lfw) \cite{LFW}, LSUN bedrooms \cite{LSUN}, and cifar-10 \cite{cifar10} datasets.

It is well-established that evaluating generative models is hard \cite{theis2016}.
Many GAN papers rely on a combination of samples for quality evaluation, supplemented by a number of heuristic quantitative measures.
We will mostly focus on training stability by showing plots of the loss function,
and will provide generated samples to claim comparable sample quality between methods, but we will avoid claiming better sample quality.
These samples are all generated at random and are not cherry-picked.

The design of $g_{\theta}$ and $\Phi_{\omega}$ are following DCGAN principles \cite{dcgan}, with both $g_{\theta}$ and $\Phi_{\omega}$ being a convolutional network with batch normalization \cite{ioffe2015batch} and ReLU activations.
$\Phi_{\omega}$ has output size $bs \times F \times 4 \times 4$. 
The inner product can then equivalently be implemented as \verb$conv(4x4, F->1)$ or \verb$flatten + Linear(4*4*F -> 1)$.
We generate $64\times64$ images for lfw and LSUN and $32\times32$ images on cifar, and train with minibatches of size 64.
We follow the experimental framework and implementation of \cite{WGAN}, where we ensure the boundedness of $\Phi_{\omega}$ by clipping the weights pointwise to the range $[-0.01, 0.01]$.

\begin{figure}
  \centering
  \includegraphics[width=0.8\linewidth]{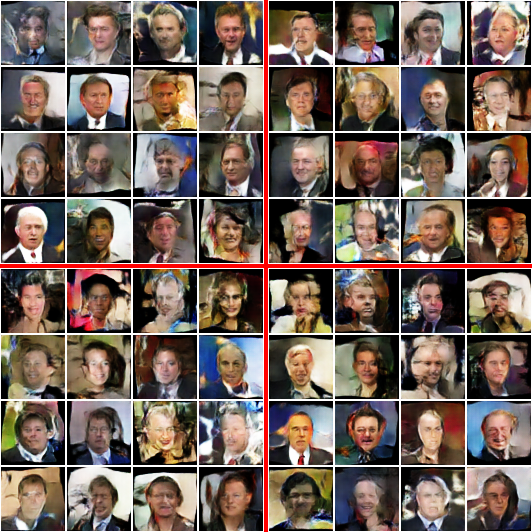}
  \vskip 0.5em
  \includegraphics[width=0.8\linewidth]{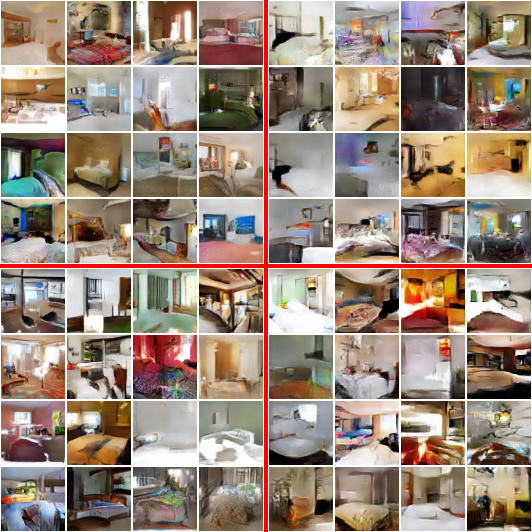}
  \caption{Samples generated with primal (left) and dual (right) formulation, in $\ell_1$ (top) and $\ell_2$ (bottom) norm. 
  (A) lfw (B) LSUN.}
  \label{fig:primaldual_lfw_lsun}
\end{figure}

\textbf{Primal versus dual form of mean matching}.
To illustrate the validity of both the primal and dual formulation,
we trained mean matching GANs both in the primal and dual form, see respectively Algorithm \ref{alg:P_mu} and \ref{alg:D_mu}.
Samples are shown in Figure \ref{fig:primaldual_lfw_lsun}.
Note that optimizing the dual form is less efficient and only feasible for mean matching, not for covariance matching.
The primal formulation of IPM$_{\mu,1}$ GAN corresponds to clipping $v$, i.e. the original WGAN, while for IPM$_{\mu,2}$ we divide $v$ by its $\ell_2$ norm if it becomes larger than 1.
In the dual, 
for $q=2$ we noticed little difference between maximizing the $\ell_2$ norm or its square. 

\begin{figure}
  \centering
  \includegraphics[width=0.8\linewidth]{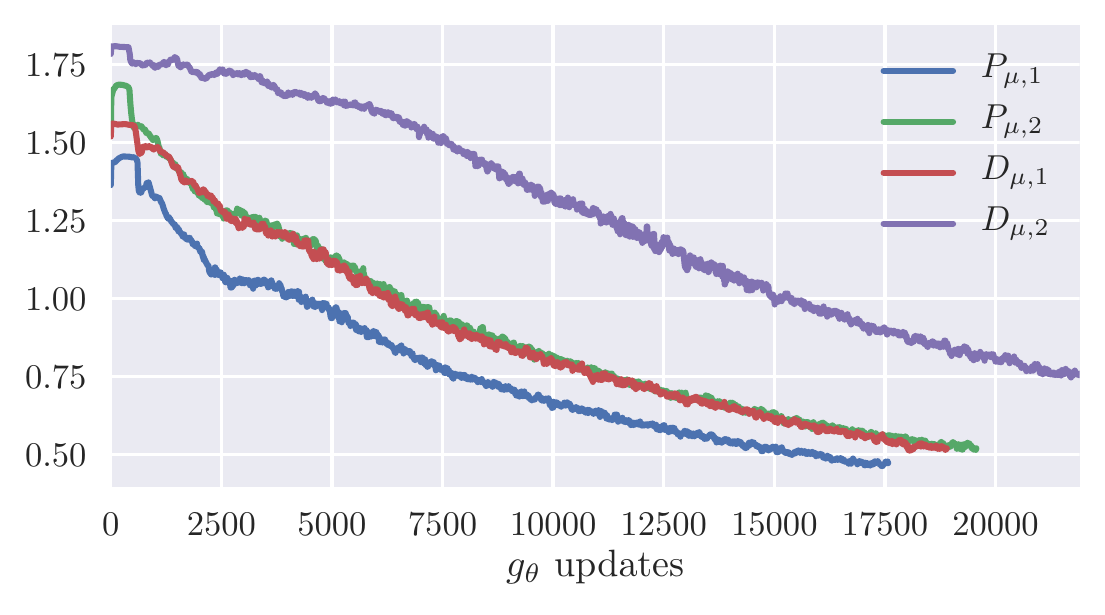}
  \caption{Plot of the loss of $P_{\mu,1}$ (i.e. WGAN), $P_{\mu,2}$ $D_{\mu,1}$ $D_{\mu,2}$ during training of lfw, as a function of number of updates to $g_\theta$.
    Similar to the observation in \cite{WGAN}, training is stable and the loss is a useful metric of progress, across the different formulations.}
  \label{fig:mean_loss}
\end{figure}

We observed that the default learning rates from WGAN (5e-5) are optimal for both primal and dual formulation.
Figure \ref{fig:mean_loss} shows the loss (i.e. IPM estimate) dropping steadily for both the primal and dual formulation independently of the choice of the $\ell_q$ norm.
We also observed that during the whole training process, samples generated from the same noise vector across iterations, remain similar in nature (face identity, bedroom style), while details and background will evolve.
This qualitative observation indicates valuable stability of the training process.

For the dual formulation (Algorithm \ref{alg:D_mu}), we confirmed the hypothesis that we need a good estimate of $\mu_{\omega}(\mathbb{P}_r)$ in order to compute the gradient of the generator $\nabla_{\theta}$: 
we needed to increase the minibatch size of real threefold to $3\times64$.

\begin{figure}[ht!]
\vskip -0.1in
  \centering
  \includegraphics[width=0.7\linewidth]{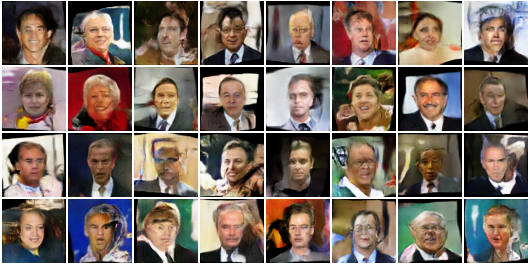}
  \includegraphics[width=0.8\linewidth]{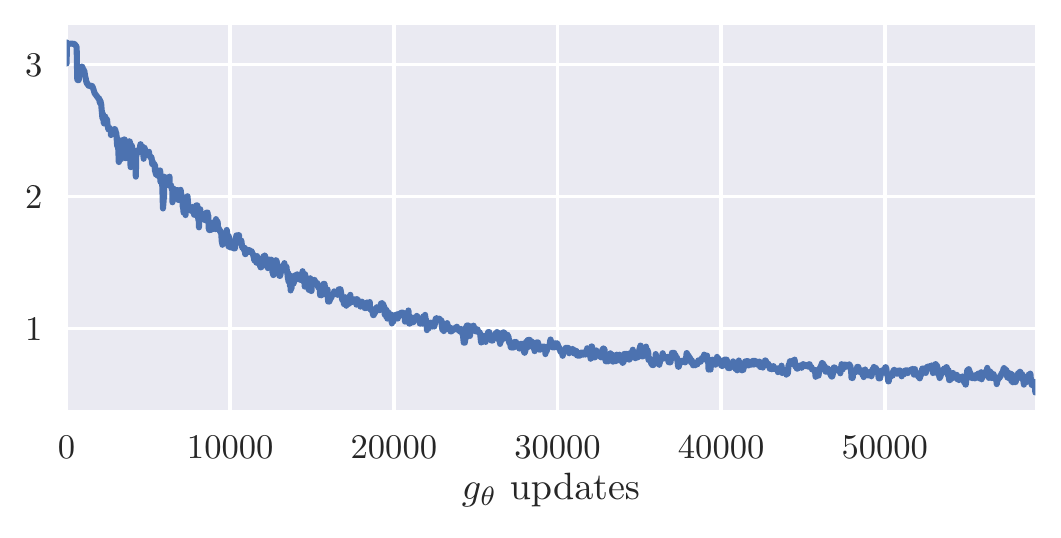}
  \vskip -0.1in
  \caption{lfw samples generated with covariance matching and plot of loss function (IPM estimate) $\hat{\mathcal{L}}_{\sigma}(U,V,\omega,\theta)$.}
  \label{fig:cov_lfw}
\end{figure}
\begin{figure}[ht!]
  \vskip -0.1in
  \centering
  \includegraphics[width=0.7\linewidth]{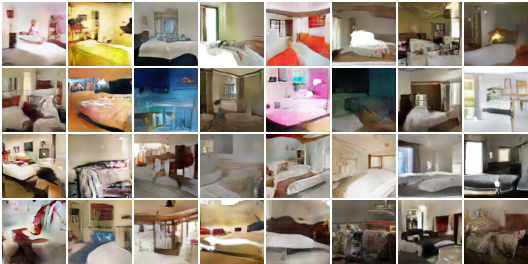}
  \includegraphics[width=0.8\linewidth]{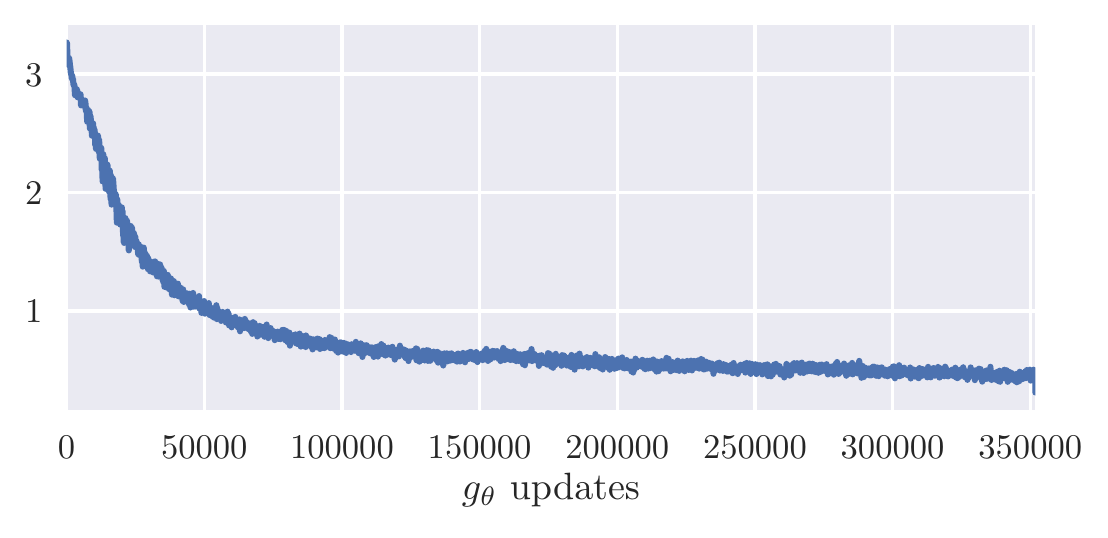}
  \vskip -0.1in
  \caption{LSUN samples generated with covariance matching and plot of loss function (IPM estimate) $\hat{\mathcal{L}}_{\sigma}(U,V,\omega,\theta)$.}
  \label{fig:cov_lsun}
  \vskip -0.2in
\end{figure}

\textbf{Covariance GAN}.
We now experimentally investigate the IPM defined by covariance matching.
For this section and the following, we use only the primal formulation, i.e. with explicit $u_j$ and $v_j$ orthonormal (Algorithm \ref{alg:P_Sigma}).
Figure \ref{fig:cov_lfw} and \ref{fig:cov_lsun} show samples and loss from lfw and LSUN training respectively.
We use Algorithm \ref{alg:P_Sigma} with $k=16$ components. 
We obtain samples of comparable quality to the mean matching formulations (Figure \ref{fig:primaldual_lfw_lsun}),
and we found training to be stable independent of hyperparameters like number of components $k$ varying between 4 and 64.

\begin{figure}[t]
  \centering
  \includegraphics[width=0.8\linewidth]{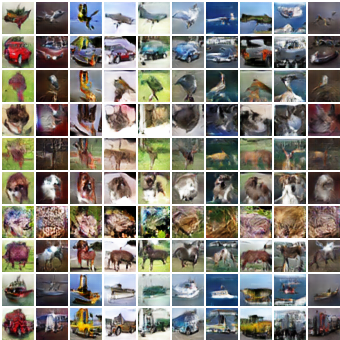}
  \caption{Cifar-10: Class-conditioned generated samples. Within each column, the random noise $z$ is shared,
    while within the rows the GAN is conditioned on the same class: from top to bottom 
    \emph{airplane, automobile, bird, cat, deer, dog, frog, horse, ship, truck}.}
  \label{fig:cifar}
\vskip -0.1in
\end{figure}

\textbf{Covariance GAN with labels and conditioning}.
\begin{table}[t]
\centering
\caption{Cifar-10: inception score of our models and baselines.}
\label{tab:inceptions}
\begin{adjustbox}{width=\linewidth}
\begin{tabular}{| l | l | l | l | }
\hline
                & Cond (+L)    & Uncond (+L)  & Uncond (-L)   \\
\hline
L1+Sigma        & 7.11 $\pm$ 0.04 & 6.93 $\pm$ 0.07 & 6.42 $\pm$ 0.09 \\
L2+Sigma        & 7.27 $\pm$ 0.04 & 6.69 $\pm$ 0.08 & 6.35 $\pm$ 0.04 \\
Sigma           & \textbf{7.29 $\pm$ 0.06} & \textbf{6.97 $\pm$ 0.10} & \textbf{6.73 $\pm$ 0.04} \\
WGAN            & 3.24 $\pm$ 0.02 & 5.21 $\pm$ 0.07 & 6.39 $\pm$ 0.07 \\
\hline
BEGAN \cite{berthelot2017began} &              &              & 5.62         \\
Impr. GAN ``-LS'' \cite{openAiGan} &              & 6.83 $\pm$ 0.06 &              \\
Impr. GAN Best  \cite{openAiGan} &              & 8.09 $\pm$ 0.07 &              \\
\hline
\end{tabular}
\end{adjustbox}
\end{table}

Finally, we conduct experiments on the cifar-10 dataset, where we will leverage the additional label information
by training a GAN with conditional generator $g_\theta(z,y)$ with label $y \in [1, K]$ supppplied as one-hot vector concatenated with noise $z$.
\nocite{conditionalGAN,infogan,acgan}
Similar to Infogan \cite{infogan} and AC-GAN \cite{acgan}, we add a new output layer, $S \in \mathbb{R}^{K\times m}$ and will write the logits $\scalT{S}{\Phi_{\omega}(x)} \in \mathbb{R}^K$.
We now optimize a combination of the IPM loss and the cross-entropy loss $CE(x,y; S, \Phi_\omega) = -\log \left[\rm{Softmax}(\scalT{S}{\Phi_{\omega}(x)})_y \right]$.
The critic loss becomes
$\mathcal{L}_D = \hat{\mathcal{L}}_{\sigma} - \lambda_D \frac{1}{N} \sum_{(x_i,y_i) \in \text{lab}} CE(x_i,y_i;S,\Phi_{\omega})$, with hyper-parameter $\lambda_D$.
We now sample three minibatches for each critic update: a labeled batch for the CE term, and for the IPM a real unlabeled + generated batch.

The generator loss (with hyper-param $\lambda_G$) becomes: 
$\mathcal{L}_G = \hat{\mathcal{L}}_{\sigma}  + \lambda_G \frac{1}{N} \sum_{z_i \sim p_z, y_i \sim p_y} CE(g_\theta(z_i,y_i),y_i;S,\Phi_{\omega})$
which still only requires a single minibatch to compute.

We confirm the improved stability and sample quality of objectives including covariance matching with inception scores \cite{openAiGan} in Table \ref{tab:inceptions}.
Samples corresponding to the best inception score (Sigma) are given in Figure \ref{fig:cifar}.
Using the code released with WGAN \cite{WGAN}, these scores come from the DCGAN model with \verb$n_extra_layers=3$ (deeper generator and discriminator) .
More samples are in appendix with combinations of Mean and Covariance Matching.
Notice rows corresponding to recognizable classes, while the noise $z$ (shared within each column) clearly determines other elements of the visual style like dominant color, across label conditioning.

\section{Discussion}
We noticed the influence of clipping on the capacity of the critic: a higher number of feature maps was needed to compensate for clipping.
The question remains what alternatives to clipping of $\Phi_{\omega}$ can ensure the boundedness.
For example, we succesfully used an $\ell_2$ penalty on the weights of $\Phi_{\omega}$. Other directions are to explore geodesic distances between the covariances \cite{log-euclideanmetrics}, and extensions of the IPM framework to the multimodal setting \cite{pix2pix}.

\bibliography{simplex}
\bibliographystyle{icml2017}

\onecolumn
\begin{center}
\textbf{Supplementary Material for McGan: Mean and Covariance Feature Matching GAN}
\\
\vskip 1em
\begin{icmlauthorlist}
\icmlauthor{Youssef  Mroueh}{equal,aif,to}
\icmlauthor{Tom Sercu}{equal,aif,to}
\icmlauthor{Vaibhava Goel}{to}
\end{icmlauthorlist}
\end{center}
\appendix
\section{Subspace Matching Interpretation of Covariance Matching GAN }
Let $\Delta_{\omega}= \Sigma_{\omega}(\mathbb{P})-\Sigma_{\omega}(\mathbb{Q})$. $\Delta_{\omega}$  is a symmetric matrix but not PSD, which has the property that its eigenvalues $\lambda_j$ are related to its singular values as given by: $\sigma_j=|\lambda_j|$ and its left and right singular vectors coincides with its eigenvectors and  satisfy the following equality $u_j=\text{ sign}(\lambda_j)v_j$.
One can ask here if we can avoid having both $U,V$ in the definition of IPM$_{\Sigma}$ since at the optimum $u_j=\pm v_j$. One could consider $\delta E_{\omega}(\mathbb{P}_r,\mathbb{P}_{\theta})$ defined as follows: 
$$\underset{\substack{\omega \in \Omega},U \in \pazocal{O}_{m,k}}{\max}~~ \underset{\substack{\text{Energy in the subspace}\\\text{of real data}}}{\underset{x\sim \mathbb{P}_r}{\mathbb{E}} \nor{U\Phi_{\omega}(x)}^2}-\underset{\substack{\text{Energy in the subspace}\\\text{of fake data}}}{ \underset{z\sim p_{z}}{\mathbb{E}}\nor{U \Phi_{\omega}(g_{\theta}(z))}^2},$$
and then solve for $\min_{g_{\theta}} \delta E_{\omega}(\mathbb{P}_r,\mathbb{P}_{\theta})$. 
Note that:
\begin{align*}
\delta E_{\omega}(\mathbb{P}_r,\mathbb{P}_{\theta}) &=\underset{\substack{\omega \in \Omega, U\in \pazocal{O}_{m,k}}}{\max} Trace(U^{\top}(\Sigma_{\omega}(\mathbb{P}_{r})-\Sigma_{\omega}(\mathbb{P}_{\theta}))U)\\
&= \max_{\omega \in \Omega}\sum_{i=1}^k \lambda_i(\Delta_{\omega})
\end{align*}
$\delta E_{\omega}$ is not symmetric furthermore the sum of those eigenvalues is not guaranteed to be positive and hence $\delta E_{\omega}$  is not guaranteed to be non negative, and hence does not define an IPM. Noting that  $\sigma_i(\Delta_{\omega})=|\lambda_{i}(\Delta_\omega)|$,we have that:
$$\text{IPM}_{\Sigma}(\mathbb{P}_r,\mathbb{P}_{\theta})=\sum_{i=1}^k \sigma_i(\Delta_{\omega})\geq \sum_{i=1}^k \lambda_i(\Delta_{\omega})=\delta E_{\omega}(\mathbb{P}_r,\mathbb{P}_{\theta}).$$
Hence $\delta E$ is not an IPM but can be optimized as a lower bound of the IPM$_{\Sigma}$. This would have an energy interpretation as in the energy based GAN introduced recently \cite{EBGAN}: the discriminator defines a  subspace that has higher energy on real data than fake data, and the generator maximizes his energy in this subspace. 
\section{Mean and Covariance Matching Loss Combinations}
\begin{figure}[ht!]
  \centering
  \includegraphics[width=\linewidth]{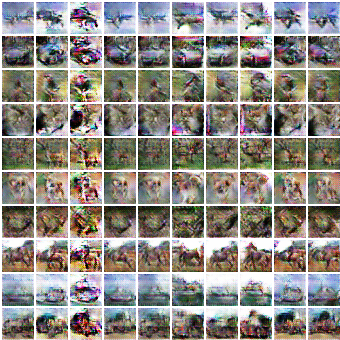}
  \caption{Cifar-10: Class-conditioned generated samples with IPM$_{\mu,1}$(WGAN). Within each column, the random noise $z$ is shared,
    while within the rows the GAN is conditioned on the same class: from top to bottom 
    \emph{airplane, automobile, bird, cat, deer, dog, frog, horse, ship, truck}.}
  \label{fig:cifarwgan}
\end{figure}

We report below samples for McGan, with different IPM$_{\mu,q}$ and IPM$_{\Sigma}$ combinations. All results are reported for the same architecture choice for generator and discriminator, which produced qualitatively good samples with IPM$_{\Sigma}$ (Same one reported in Section 6 in the main paper).
Note that in Figure \ref{fig:cifarwgan} with the same hyper-parameters and architecture choice, WGAN failed to produce good sample. In other configurations training converged.

\begin{figure}[ht!]
  \centering
  \includegraphics[width=\linewidth]{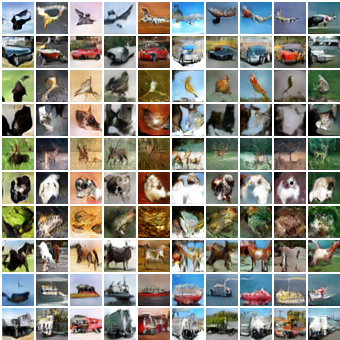}
  \caption{Cifar-10: Class-conditioned generated samples with IPM$_{\mu,2}$. Within each column, the random noise $z$ is shared,
    while within the rows the GAN is conditioned on the same class: from top to bottom 
    \emph{airplane, automobile, bird, cat, deer, dog, frog, horse, ship, truck}.}
  \label{fig:cifarl2}
\end{figure}

\begin{figure}
  \centering
  \includegraphics[width=\linewidth]{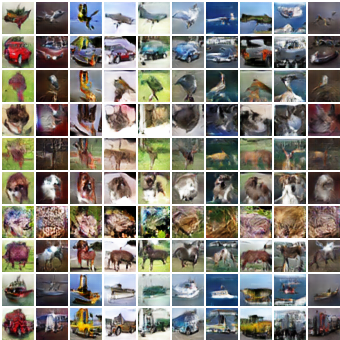}
  \caption{Cifar-10: Class-conditioned generated samples with IPM$_{\Sigma}$. Within each column, the random noise $z$ is shared,
    while within the rows the GAN is conditioned on the same class: from top to bottom 
    \emph{airplane, automobile, bird, cat, deer, dog, frog, horse, ship, truck}.}
  \label{fig:cifarSigma}
\end{figure}

\begin{figure}
  \centering
  \includegraphics[width=\linewidth]{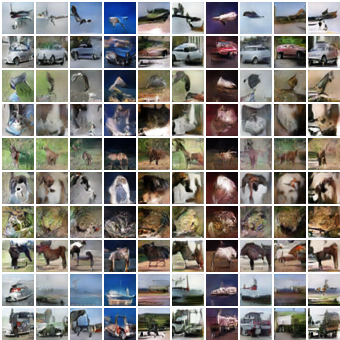}
  \caption{Cifar-10: Class-conditioned generated samples with IPM$_{\mu,1}$+ IPM$_{\Sigma}$. Within each column, the random noise $z$ is shared,
    while within the rows the GAN is conditioned on the same class: from top to bottom 
    \emph{airplane, automobile, bird, cat, deer, dog, frog, horse, ship, truck}.}
  \label{fig:cifarWGANSigma}
\end{figure}

\begin{figure}
  \centering
  \includegraphics[width=\linewidth]{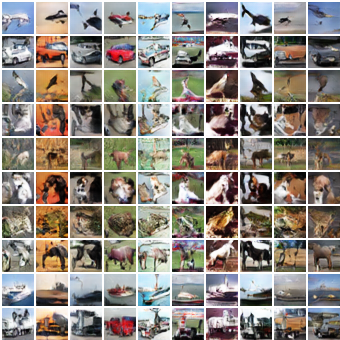}
  \caption{Cifar-10: Class-conditioned generated samples with IPM$_{\mu,2}$+ IPM$_{\Sigma}$. Within each column, the random noise $z$ is shared,
    while within the rows the GAN is conditioned on the same class: from top to bottom 
    \emph{airplane, automobile, bird, cat, deer, dog, frog, horse, ship, truck}.}
  \label{fig:cifarL2Sigma}
\end{figure}

\end{document}